\documentclass{article}

\usepackage[preprint]{neurips_2025}

\usepackage[utf8]{inputenc} 
\usepackage[T1]{fontenc}    
\usepackage{hyperref}       
\usepackage{url}            
\usepackage{booktabs}       
\usepackage{amsfonts}       
\usepackage{nicefrac}       
\usepackage{microtype}      
\usepackage{xcolor}         
\usepackage{graphicx}       
\usepackage{float}
\graphicspath{{media/}}     
\usepackage{multirow}       
\usepackage{tabularx}
\usepackage{amsmath}
\usepackage{subcaption}

\title{Few-Shot Semantic Segmentation Meets SAM3}

\author{%
  Yi-Jen Tsai$^{1}$ \quad Yen-Yu Lin$^{1}$ \quad Chien-Yao Wang$^{2}$\thanks{Corresponding author} \\
  $^{1}$National Yang Ming Chiao Tung University \\
  $^{2}$Institute of Information Science, Academia Sinica, Taiwan \\
  \texttt{tsai.cs14@nycu.edu.tw, lin@cs.nycu.edu.tw, kinyiu@iis.sinica.edu.tw}
}

\begin{document}

\maketitle

\begin{abstract}
Few-Shot Semantic Segmentation (FSS) focuses on segmenting novel object categories from only a handful of annotated examples. Most existing approaches rely on extensive episodic training to learn transferable representations, which is both computationally demanding and sensitive to distribution shifts. In this work, we revisit FSS from the perspective of modern vision foundation models and explore the potential of Segment Anything Model 3 (SAM3) as a training-free solution. By repurposing its Promptable Concept Segmentation (PCS) capability, we adopt a simple spatial concatenation strategy that places support and query images into a shared canvas, allowing a fully frozen SAM3 to perform segmentation without any fine-tuning or architectural changes. Experiments on PASCAL-$5^i$ and COCO-$20^i$ show that this minimal design already achieves state-of-the-art performance, outperforming many heavily engineered methods. Beyond empirical gains, we uncover that negative prompts can be counterproductive in few-shot settings, where they often weaken target representations and lead to prediction collapse despite their intended role in suppressing distractors. These findings suggest that strong cross-image reasoning can emerge from simple spatial formulations, while also highlighting limitations in how current foundation models handle conflicting prompt signals. Code at: \url{https://github.com/WongKinYiu/FSS-SAM3}
\end{abstract}

\section{Introduction}
Semantic segmentation is a fundamental task in computer vision, aiming to assign a semantic label to every pixel in an image. Despite remarkable progress driven by deep learning, current approaches still rely heavily on large-scale, densely annotated datasets. However, acquiring such annotations is costly and often impractical, especially in domains like medical imaging or industrial inspection. Moreover, models trained on fixed category sets tend to struggle when encountering unseen objects, limiting their applicability in open-world scenarios.

To address these challenges, Few-Shot Semantic Segmentation (FSS) has emerged as a promising approach. Instead of requiring extensive labeled data, FSS aims to segment novel object categories given only a few annotated examples, referred to as the support set. Most existing methods follow a similarity-based paradigm, where query regions are identified by matching features extracted from the support images. This design enables knowledge transfer across categories, but still depends on carefully learned representations.

In practice, such representations are obtained through large-scale episodic training, where models are trained on simulated few-shot tasks constructed from base classes. While effective, this paradigm is computationally expensive and often sensitive to distribution shifts between training and evaluation categories. At the same time, recent advances in visual foundation models, such as CLIP, DINO, and Segment Anything (SAM), have demonstrated strong generalization across diverse visual tasks. In particular, SAM introduces promptable segmentation, allowing flexible interaction through points, boxes, or masks, and provides a viable alternative to traditional FSS, enabling flexible segmentation without dedicated training.

As the latest member of the Segment Anything family, SAM3 extends the capabilities of its predecessors by introducing Promptable Concept Segmentation (PCS), enabling segmentation based on textual descriptions, visual exemplars, or both. Unlike earlier versions that focus on class-agnostic segmentation, SAM3 is inherently designed to operate at the level of semantic concepts. This raises a natural question: can few-shot segmentation be achieved by directly leveraging such a pretrained model, without any additional training?

In this work, we explore a fully frozen SAM3 for FSS. Instead of introducing new modules or adapting the model, we adopt a simple spatial concatenation strategy that places support and query images into a shared canvas. Under this formulation, cross-image correspondence emerges naturally from the model’s attention mechanism, allowing SAM3 to perform segmentation in a single forward pass. Despite its simplicity, this approach achieves strong performance on standard benchmarks such as COCO-$20^i$ and PASCAL-$5^i$, outperforming many training-based methods.

Beyond performance, our study reveals an unexpected phenomenon: negative prompts consistently degrade performance in few-shot settings. Contrary to the common intuition that negative guidance suppresses distractors, our analysis reveals that such prompts can paradoxically introduce semantic ambiguity, triggering a collapse in prediction quality. This finding highlights a limitation in how current foundation models handle competing semantic signals and suggests the need for more robust prompt interaction mechanisms.

The main contributions of this work are summarized as follows:
\begin{enumerate}
    \item \textbf{Training-Free Few-Shot Segmentation via Spatial Concatenation} 
    We propose a simple yet effective framework that reformulates few-shot segmentation as a spatial reasoning problem by concatenating support and query images into a shared canvas, enabling cross-image correspondence without any training.
    \item \textbf{A Strong Frozen SAM3 Baseline} 
    We demonstrate that a fully frozen SAM3, combined with the proposed formulation, achieves competitive performance on standard benchmarks such as COCO-$20^i$ and PASCAL-$5^i$, outperforming many training-based approaches.
    \item \textbf{Empirical Analysis of Negative Prompt Failure} 
    We systematically investigate the effect of negative prompts and reveal a consistent performance degradation, providing new insights into prompt interactions in foundation models.
\end{enumerate}

\section{Related Work}
\subsection{Classical Few-Shot Segmentation}
Few-Shot Semantic Segmentation (FSS) aims to segment target objects in a query image given only a few annotated support examples. Existing methods can be broadly categorized into two types based on how support information is utilized.
\paragraph{Prototypical Matching}
Early FSS methods, such as PFENet \cite{tian2020prior}, BAM \cite{lang2022learning}, SSP \cite{fan2022self}, and FPTrans \cite{zhang2022feature}, primarily focus on feature compression. These methods typically employ Mask Average Pooling to condense support features into representative global prototypes, and then perform segmentation by measuring the similarity between these prototypes and query features. While this strategy reduces computational cost and provides a compact representation, it inevitably discards spatial details that are often critical for distinguishing objects with complex structures or large appearance variations. As a result, these methods tend to struggle when fine-grained correspondence is required.
\paragraph{Pixel-wise Matching}
To reduce the information loss caused by the use of prototype representations, subsequent works, including HSNet\cite{min2021hypercorrelation}, SCCAN\cite{xu2023self}, and HDMNet\cite{peng2023hierarchical} directly match pixels between support and query features. By using techniques such as cross-attention or 4D convolutions, these models can capture more detailed relationships. While they achieve strong results on benchmarks, these approaches heavily rely on large-scale episodic training constructed from base classes. As a result, their performance is often sensitive to shifts in the distribution between base and novel classes.

\subsection{Foundation Models for Segmentation}
The emergence of Visual Foundation Models (VFMs), such as CLIP \cite{radford2021learning}, DINO \cite{caron2021emerging}, and SAM \cite{kirillov2023segment}, has introduced a new paradigm for segmentation by providing general-purpose visual representations pretrained on large-scale data. Specifically, the Segment Anything Model (SAM) series has become a cornerstone for FSS, due to its zero-shot capacity and flexible prompting mechanisms.
\paragraph{SAM and Prompting Capabilities}
The Segment Anything Model (SAM) \cite{kirillov2023segment} demonstrates strong zero-shot, class-agnostic segmentation performance and supports interactive segmentation via various prompts such as points, boxes, and masks. Expanding on its capabilities, several studies have adapted SAM for FSS by incorporating additional modules to establish support–query correspondence, followed by prompt-based segmentation. For instance, VRP-SAM \cite{sun2024vrp} replaces the original prompt encoder with a specialized visual reference prompt encoder, while GF-SAM \cite{zhang2024bridge} utilizes external encoders (e.g., DINOv2 \cite{oquab2023dinov2}) to generate prior masks as guidance. Despite their effectiveness, these approaches often require auxiliary image encoders or complex prompt-generation pipelines.
\paragraph{SAM2 and Memory Mechanism} 
SAM2 \cite{ravi2024sam} extends promptable segmentation to the video domain by introducing a memory attention mechanism that enables feature matching across frames. It follows a prompt-and-propagate paradigm, where objects in the initial frame are specified via points, boxes, or masks and subsequently tracked across frames based on visual similarity, enabling mask propagation over time with high spatial precision. Methods such as SANSA \cite{cuttano2025sansa} and FSSAM \cite{xu2025unlocking} adapt this design to few-shot segmentation by treating support images as prior frames stored in memory, effectively shifting the task from object tracking to semantic tracking based on shared concepts across images. However, this memory mechanism is primarily designed for temporal consistency rather than semantic alignment across different instances, and thus often requires additional training or adaptation modules to bridge this gap.
\paragraph{SAM3 and Concept Segmentation} 
As the latest iteration of the SAM series, SAM3 \cite{carion2025sam} further advances the SAM framework by introducing Promptable Concept Segmentation (PCS), enabling segmentation based on semantic concepts expressed through text or visual exemplars. Unlike SAM and SAM2, which primarily focus on class-agnostic objectness, SAM3 is more naturally aligned with the objectives of FSS, as it supports concept-level reasoning across different instances. This shift suggests that few-shot segmentation may be addressed more directly, without relying heavily on fine-tuning or auxiliary modules that were used in previous works to bridge the semantic gap. Motivated by this observation, we revisit Few-Shot Semantic Segmentation from a different perspective and investigate whether cross-image correspondence can be achieved through a simple spatial formulation within a shared canvas.

\subsection{Prompt Engineering and Visual Prompting}
As segmentation models become increasingly prompt-driven, understanding how to provide precise guidance for these models has become an important research direction. PerSAM \cite{zhang2023personalize} introduces the concept of personalized segmentation, automatically generating positive prompts through feature matching to achieve one-shot adaptation. More recently, SSPrompt \cite{huang2024learning} identifies a semantic gap in the original SAM; since it was pretrained primarily for class-agnostic segmentation, the model struggles to resolve high-level categories without explicit spatial guidance. To bridge this, they introduce learnable semantic embeddings to ground the model’s predictions. In contrast, our approach leverages the native Promptable Concept Segmentation (PCS) of SAM3, which—unlike earlier versions—is inherently optimized for semantic understanding without additional prompt learning.

While these studies optimize the efficiency of positive guidance, the role of negative prompts remains under-explored, particularly in few-shot settings where semantic ambiguity is more pronounced. Our work is related to these directions in two aspects. First, we demonstrate that a fully frozen SAM3, combined with a simple spatial concatenation strategy, can serve as a strong baseline without additional adapters or complex pipelines. Second, we examine the role of negative prompts, which, contrary to common assumptions, may disrupt semantic alignment and degrade performance. These observations suggest that prompt interactions in foundation models are not yet fully understood.

\begin{figure}[t]
    \centering
    \includegraphics[width=0.9\textwidth]{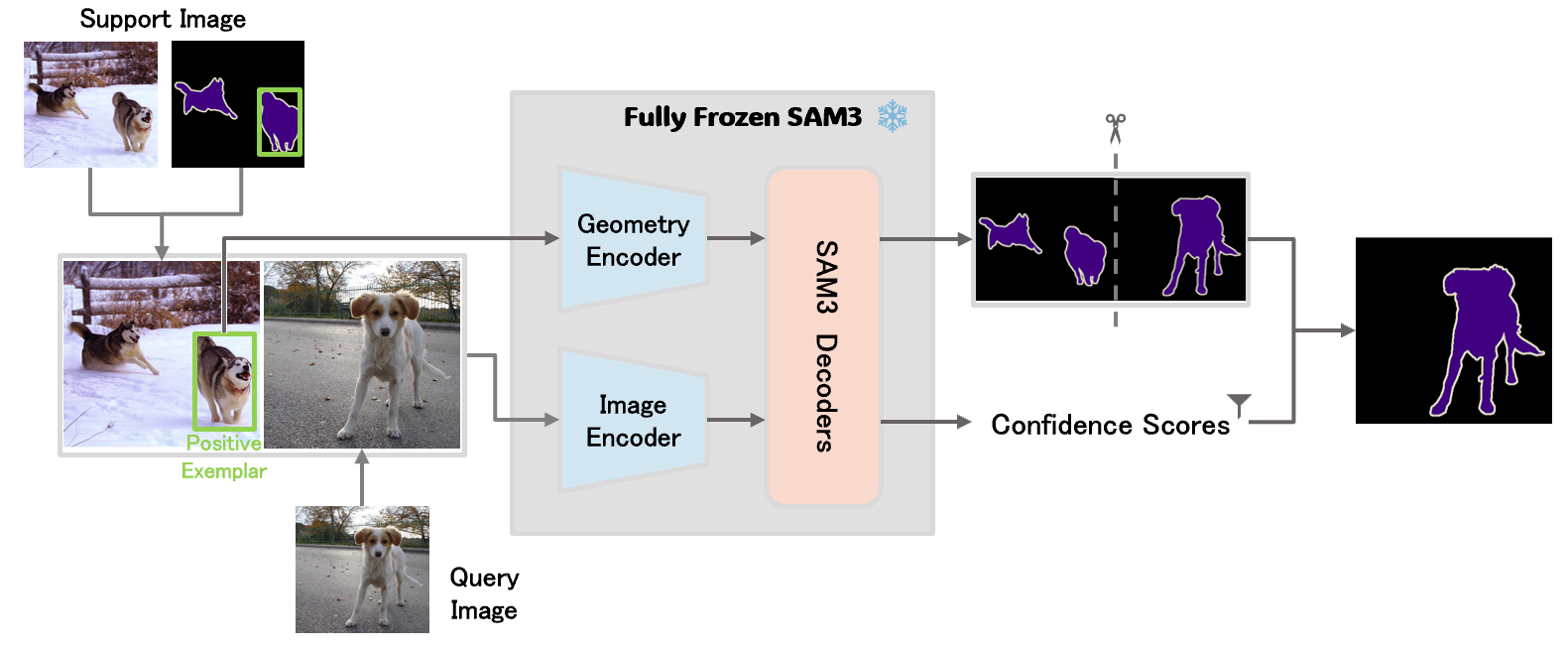}
    \caption{\textbf{Pipeline of our SAM3-based FSS framework}. By combining instance-aware positive prompts with a unified spatial formulation, our method enables the fully frozen SAM3 to perform implicit cross-image feature matching in a single forward pass without architectural modifications.} 
    \label{fig:framework}
\end{figure}

\section{Methodology}
In this work, we investigate how the large vision foundation model, SAM3, can be adapted to the Few-Shot Segmentation (FSS) setting without any additional training, fine-tuning, or architectural modification. Although SAM3 demonstrates strong zero-shot segmentation capability, it lacks an explicit mechanism for cross-image correspondence, which is a fundamental component of conventional FSS pipelines. To address this, we introduce a unified spatial formulation that enables support and query interaction within a single forward pass, together with instance-aware prompting designed to improve spatial precision.

\subsection{Problem Definition}
Few-shot segmentation aims to segment objects from novel categories using a limited number of annotated support examples. Formally, given a support set $S = \{(I_s^i, M_s^i)\}_{i=1}^K$, where $I_s^i$ denotes the support image and $M_s^i$ its binary mask, the goal is to predict the corresponding object mask $M_q$ for a query image $I_q$. Following standard FSS evaluation protocols, inference is performed episodically. For each dataset fold, we sample $1000$ independent test episodes where support and query images are different samples ($I_s \neq I_q$). This setting evaluates the model’s ability to generalize to unseen categories under limited supervision.

\subsection{Framework Overview}
Our framework adopts a fully frozen SAM3 model as the segmentation backbone. Unlike conventional FSS methods that rely on dual-branch architectures and explicit feature matching networks, we enable cross-image reasoning directly within the attention mechanism of the pretrained SAM3. Our key idea is to reformulate the FSS task as a spatial reasoning problem within a shared coordinate system by embedding both support and query images into a unified spatial canvas. Under this formulation, tokens from both images participate in the same attention computation, allowing implicit cross-image interaction without modifying the pretrained model. As illustrated in Figure \ref{fig:framework}, the overall pipeline consists of three stages: (1) instance-aware prompt extraction from the support annotation; (2) spatial composition of support and query images into a unified canvas; and (3) cross-image segmentation using the
frozen SAM3 decoder guided by geometric prompts to implicitly align features and predict the final target mask. Together, these components enable SAM3 to perform few-shot segmentation using only visual exemplars while keeping the entire model unchanged.

\subsection{Unified Spatial Canvas}
Conventional FSS approaches typically rely on explicit support–query matching modules, implemented by dual encoders and feature fusion mechanisms. Such architectures are specifically designed for cross-image correspondence but require additional training and architectural modifications. In contrast, our approach enables cross-image interaction through a spatial composition strategy that preserves the original architecture of SAM3. Specifically, the support and query images are geometrically aligned into a shared global coordinate frame, forming a composite canvas that is jointly processed by SAM3, $I_{\text{canvas}} = \text{Compose}(I_s, I_q)$. Geometric prompts derived from the support annotations are correspondingly transformed into the coordinate system of the global canvas. Under this formulation, tokens originating from both images are processed jointly by the self-attention layers of SAM3. As a result, the model can implicitly associate semantically related regions across images through its native attention mechanism. This design effectively converts the cross-image matching problem into a spatial reasoning task within a unified attention space, allowing SAM3 to perform few-shot segmentation without introducing additional modules.

\subsection{Instance-Aware Prompting} 
We observed that foundation segmentation models like SAM3 are highly sensitive to the spatial precision of geometric prompts. In standard FSS pipelines, prompts are typically derived from semantic masks, which may contain multiple object instances or large background regions. Such coarse prompts can introduce ambiguity and reduce segmentation accuracy. To improve prompt quality, we introduce an instance-aware prompting strategy that derives geometric prompts from individual object instances rather than the entire semantic mask.
\paragraph{Positive exemplar formulation}
Instead of generating prompts from the entire semantic mask, we exploit instance-level annotations. From the support image, we select a representative object instance and derive a tight geometric prompt from its spatial extent. This instance-level prompt provides a localized and unambiguous reference to the target object. By focusing on a single representative instance, the model can better capture discriminative appearance features while avoiding interference from irrelevant background regions.
\paragraph{Negative exemplar generation}
In addition to positive prompts derived from the support annotation, we explore whether automatically generated negative exemplars can further improve segmentation robustness. The intuition is that identifying potential background or irrelevant regions in the support image may help the model suppress visually similar distractors when segmenting the query image. To obtain such information, we perform an auxiliary segmentation step on the support image. Specifically, the support image together with the positive bounding-box prompt is first fed into SAM3 to produce a set of candidate mask predictions with associated confidence scores $\{(m_i, s_i)\}$. Based on a confidence threshold $\tau$, we separate the predictions into two groups:\[\mathcal{P}^{+} = \{ m_i \mid s_i \ge \tau \}, \quad\mathcal{P}^{-} = \{ m_i \mid s_i < \tau \}.\] Predictions with confidence scores higher than a predefined threshold are treated as additional positive exemplars, while predictions with lower confidence scores are regarded as potential distractor regions and used as negative exemplars. These positive and negative exemplars are then converted into geometric prompts. Following the unified spatial canvas formulation, the support and query images are composed into a single canvas, and both types of prompts are provided to SAM3 during inference. In this way, the model receives explicit guidance about both target regions and potential distractors in the support image. The empirical behavior of this negative prompting strategy is analyzed in Section~\ref{sec:neg_exp}.

\subsection{Multimodal Integration with Textual Priors} 
Visual exemplars provide strong appearance cues but may sometimes bias the model toward the specific pose, lighting conditions, or precise localization of the support instance. To improve category-level robustness, we further introduce textual category priors as an additional source of semantic information. Specifically, for each FSS episode, the class name of the target category is processed by the frozen text encoder of SAM3 to produce a semantic concept embedding. This embedding is integrated with the visual and geometric prompt representations in the SAM3 decoder. By combining visual exemplars with textual category descriptions, our framework benefits from both the discriminative appearance guidance of the visual exemplar and the semantic robustness of the textual prior. This multimodal integration encourages a broader category-level focus rather than instance-specific details, improving generalization across diverse query images.

\begin{table}[t]
  \caption{Comparison with state-of-the-art methods on PASCAL-$5^{i}$ dataset.}
  \label{tab:results_pascal}
  \centering
  \small
  \setlength{\tabcolsep}{2.5pt} 
  \begin{tabular}{l cccccc cccccc}
    \toprule
    \multirow{2}{*}{Method} & \multicolumn{6}{c}{1-shot} & \multicolumn{6}{c}{5-shot} \\
    \cmidrule(r){2-7} \cmidrule(l){8-13}
    & $5^0$ & $5^1$ & $5^2$ & $5^3$ & Mean & FB-IoU & $5^0$ & $5^1$ & $5^2$ & $5^3$ & Mean & FB-IoU \\
    \midrule
    \addlinespace[2pt]
    \multicolumn{13}{l}{\textit{\textbf{Classical FSS Methods}}} \\
    \addlinespace[3pt]
    PFENet (TPAMI'20) \cite{tian2020prior} & 61.7 & 69.5 & 55.4 & 56.3 & 60.8 & -- & 63.1 & 70.7 & 55.8 & 57.9 & 61.9 & -- \\
    HSNet (ICCV'21) \cite{min2021hypercorrelation} & 64.3 & 70.7 & 60.3 & 60.5 & 64.0 & -- & 70.3 & 73.2 & 67.4 & 67.1 & 69.5 & -- \\
    BAM (CVPR'22) \cite{lang2022learning} & 69.0 & 73.6 & 67.6 & 61.1 & 67.8 & 79.7 & 70.6 & 75.1 & 70.8 & 67.2 & 70.9 & 82.2 \\
    SCCAN (ICCV'23) \cite{xu2023self} & 68.3 & 72.5 & 66.8 & 59.8 & 66.8 & 77.7 & 72.3 & 74.1 & 69.1 & 65.6 & 70.3 & 81.8 \\
    FPTrans (NeurIPS'22) \cite{zhang2022feature} & 72.3 & 70.6 & 68.3 & 64.1 & 68.8 & -- & 76.7 & 79.0 & 81.0 & 75.1 & 78.0 & -- \\
    HDMNet (CVPR'23) \cite{peng2023hierarchical} & 71.0 & 75.4 & 68.9 & 62.1 & 69.4 & -- & 71.3 & 76.2 & 71.3 & 68.5 & 71.8 & -- \\
    PAM (AAAI'24) \cite{wang2024adaptive} & 71.1 & 75.5 & 67.0 & 64.5 & 69.5 & -- & 74.7 & 78.0 & 75.3 & 70.8 & 74.7 & -- \\
    AENet (ECCV'24) \cite{xu2024eliminating} & 72.2 & 75.5 & 68.5 & 63.1 & 69.8 & 80.8 & 74.2 & 76.5 & 74.8 & 70.6 & 74.1 & 84.5 \\
    AMNet (NeurIPS'23) \cite{wang2023focus} & 71.1 & 75.9 & 69.7 & 63.7 & 70.1 & -- & 73.2 & 77.8 & 73.2 & 68.7 & 73.2 & -- \\
    HMNet (NeurIPS'24) \cite{xu2024hybrid} & 72.2 & 75.4 & 70.0 & 63.9 & 70.4 & 81.6 & 74.0 & 77.2 & 74.1 & 70.5 & 73.9 & 84.4 \\
    \midrule
    \addlinespace[2pt]
    \multicolumn{13}{l}{\textit{\textbf{Foundation-based FSS Methods}}} \\ 
    \addlinespace[3pt]
    Matcher (ICLR'24) \cite{liu2023matcher} & 67.7 & 70.7 & 66.9 & 67.0 & 68.1 & -- & 71.4 & 77.5 & 74.1 & 72.8 & 74.0 & -- \\
    VRP-SAM (CVPR'24) \cite{sun2024vrp} & 73.9 & 78.3 & 70.6 & 65.1 & 71.9 & -- & -- & -- & -- & -- & -- & -- \\
    GF-SAM (NeurIPS'24) \cite{zhang2024bridge} & 71.1 & 75.7 & 69.2 & 73.3 & 72.1 & -- & \underline{81.5} & 86.3 & 79.7 & \underline{82.9} & \underline{82.6} & -- \\
    FSSAM (ICML'25) \cite{xu2025unlocking} & \textbf{81.6} & \underline{84.9} & \textbf{81.6} & 76.0 & \underline{81.0} & \underline{89.4} & \textbf{84.1} & \textbf{88.5} & \textbf{83.8} & \textbf{85.0} & \textbf{85.4} & \textbf{91.9} \\
    \midrule
    \addlinespace[2pt]
    \textbf{FSS-SAM3 (Ours)} & 70.4 & 84.6 & \underline{80.9} & \textbf{82.3} & 79.6 & 88.3 & 74.8 & 86.8 & 80.4 & 82.9 & 81.2 & 88.9 \\
    \textbf{FSS-SAM3 + text (Ours)} & \underline{76.2} & \textbf{87.5} & \textbf{81.6} & \underline{79.5} & \textbf{81.2} & \textbf{89.4} & 76.2 & \underline{86.7} & \underline{80.3} & 81.9 & 81.3 & \underline{89.0} \\
    \bottomrule
  \end{tabular}
\end{table}

\section{Experiments}

\subsection{Experiment Setup}
\paragraph{Datasets}
To evaluate the proposed method, we conduct experiments on two widely used Few-Shot Semantic Segmentation (FSS) benchmarks: PASCAL-$5^i$ and COCO-$20^i$. The PASCAL-$5^i$ dataset is built upon PASCAL VOC 2012 with additional annotations from the SDS dataset, comprising 20 object categories. Following standard protocols, the 20 classes are divided into four disjoint folds (five classes per fold) for cross-validation. COCO-$20^i$, derived from the MSCOCO dataset, contains 80 classes divided into four folds, with 20 classes each. Compared to PASCAL-$5^i$, COCO-$20^i$ is more challenging due to its greater variety of objects, significant scale variations, and complex backgrounds. Following common practice in FSS, we evaluate our model using a cross-validation approach. For each fold, we randomly sample 1,000 episodes to report the final results, ensuring statistical stability in our performance evaluation.
\paragraph{Evaluation Metrics}
Two standard metrics, mean intersection over union (mIoU) and foreground-background IoU (FB-IoU), are utilized to measure the segmentation performance. The mIoU is calculated by averaging the IoU values across all test classes within a fold, serving as the primary metric for assessing the model's accuracy on specific categories. The FB-IoU metric treats all target categories as the foreground and the remaining regions as background, then calculates the average IoU. FB-IoU is particularly useful for evaluating the model's ability to generalize across foreground and background classes, thereby reducing the impact of class imbalance during testing.
\paragraph{Implementation details}
Our framework is built upon the official implementation of SAM3. The entire model remains fully frozen, with no additional training or fine-tuning. For each episode, support and query images are concatenated into a unified spatial canvas to enable cross-image interaction. The canvas is then resized to a fixed resolution of $1008 \times 1008$, following the processing pipeline of SAM3. The geometric prompts derived from the support annotation are correspondingly transformed into the coordinate system of the composite canvas. Specifically, the instance-level bounding box is used as the positive geometric prompt; negative prompts are not used in the inference pipeline due to performance degradation. When textual guidance is enabled, category names are processed by the frozen text encoder to produce semantic embeddings, which are then fused with visual prompts in the decoder. Final query masks are predicted in a single forward pass without iterative refinement, and results are averaged across all evaluation episodes.

\subsection{Few Shot Segmentation Performance}

\begin{figure}[t]
    \centering
    \includegraphics[width=0.9\textwidth]{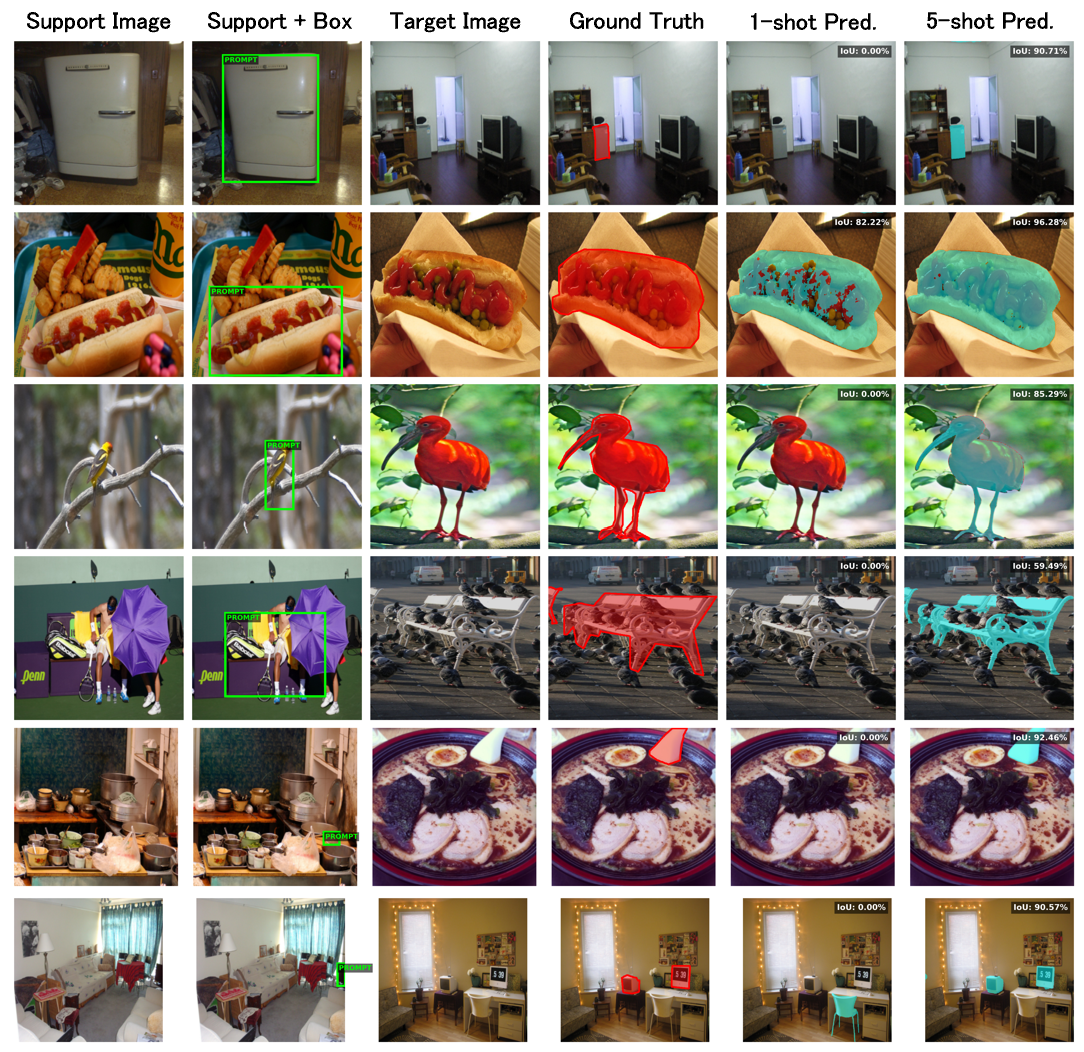} 
    \caption{visualization of 1-shot/5-shot prediction on PASCAL-$5^{i}$ dataset.}
    \label{fig:fssvis}
\end{figure}

We evaluate our method on PASCAL-$5^i$ and COCO-$20^i$ benchmarks under standard Few-Shot Segmentation (FSS) evaluation in 1-shot and 5-shot settings. The support and query images are composed into a unified spatial canvas, where box prompts from the support image serve as positive exemplars. Tables \ref{tab:results_pascal} and \ref{tab:results_coco} present comparisons with existing FSS methods. 

\paragraph{Results on PASCAL-$5^i$} On PASCAL-$5^i$, our method achieves state-of-the-art performance in the 1-shot setting. Specifically, FSS-SAM3 + text attains 81.6 mIoU, outperforming all prior methods. In the 5-shot setting, our method achieves 81.3 mIoU, which is competitive with recent foundation-based approaches but does not outperform the best-performing method. We observe that the performance gain from 1-shot to 5-shot is relatively limited in our framework. Compared to prior methods that explicitly aggregate multiple support examples, our approach adopts a simpler formulation without dedicated multi-shot fusion, which may limit the benefit from additional support images. A similar trend is observed for textual priors, whose contribution becomes marginal in the 5-shot setting, where visual exemplars already provide strong guidance for PASCAL-$5^i$ dataset.
\paragraph{Results on COCO-$20^i$} On the more challenging COCO-$20^i$ benchmark, our method consistently outperforms all state-of-the-art approaches under both settings. Notably, even without textual priors, FSS-SAM3 achieves 66.1 mIoU in the 1-shot setting, surpassing the previous best result (62.3 mIoU). When incorporating textual prompts, performance improves further to 75.4 mIoU, yielding a substantial gain of +13.1 over prior work. A similar trend is observed in the 5-shot setting, where our method achieves 71.2 mIoU without text and 74.5 mIoU with text, both exceeding the performance of existing methods. These results demonstrate that our spatial canvas formulation alone is highly effective, while textual priors provide an additional and significant performance boost.

\paragraph{Effect of textual priors} We observe that incorporating textual priors consistently improves performance across both datasets. The improvement is particularly pronounced on COCO-$20^i$, where performance increases from 66.1 to 75.4 mIoU (+9.3). This indicates that textual guidance plays a crucial role in enhancing category-level generalization under large-scale category diversity and complex visual variations.

Overall, these results demonstrate that the proposed unified spatial canvas formulation effectively enables cross-image reasoning within SAM3, allowing the model to perform few-shot segmentation without any architectural modification. The strong performance across both benchmarks indicates that simple spatial composition, when combined with appropriate prompting, is sufficient to bridge support and query images within a single attention space.

\begin{table}[t]
  \caption{Comparison with state-of-the-art methods on COCO-$20^{i}$ dataset.}
  \label{tab:results_coco}
  \centering
  \small
  \setlength{\tabcolsep}{2.4pt}
  \begin{tabular}{l cccccc cccccc}
    \toprule
    \multirow{2}{*}{Method} & \multicolumn{6}{c}{1-shot} & \multicolumn{6}{c}{5-shot} \\
    \cmidrule(r){2-7} \cmidrule(l){8-13}
    & $20^0$ & $20^1$ & $20^2$ & $20^3$ & Mean & FB-IoU & $20^0$ & $20^1$ & $20^2$ & $20^3$ & Mean & FB-IoU \\
    \midrule
    \addlinespace[2pt]
    \multicolumn{13}{l}{\textit{\textbf{Classical FSS Methods}}} \\
    \addlinespace[3pt]
    PFENet (TPAMI'20) \cite{tian2020prior} & 36.5 & 38.6 & 34.5 & 33.8 & 35.8 & -- & 36.5 & 43.3 & 37.8 & 38.4 & 39.0 & -- \\
    HSNet (ICCV'21) \cite{min2021hypercorrelation} & 36.3 & 43.1 & 38.7 & 38.7 & 39.2 & -- & 43.3 & 51.3 & 48.2 & 45.0 & 46.9 & -- \\
    BAM (CVPR'22) \cite{lang2022learning} & 43.4 & 50.6 & 47.5 & 43.4 & 46.2 & -- & 49.3 & 54.2 & 51.6 & 49.6 & 51.2 & -- \\
    SCCAN (ICCV'23) \cite{xu2023self} & 40.4 & 49.7 & 49.6 & 45.6 & 46.3 & 69.9 & 47.2 & 57.2 & 59.2 & 52.1 & 53.9 & 74.2 \\
    FPTrans (NeurIPS'22) \cite{zhang2022feature} & 44.4 & 48.9 & 50.6 & 44.0 & 47.0 & -- & 54.2 & 62.5 & 61.3 & 57.6 & 58.9 & -- \\
    PAM (AAAI'24) \cite{wang2024adaptive} & 44.1 & 55.0 & 46.5 & 48.5 & 48.5 & -- & 48.1 & 60.8 & 54.8 & 51.9 & 53.9 & -- \\
    AENet (ECCV'24) \cite{xu2024eliminating} & 43.1 & 56.0 & 50.3 & 48.4 & 49.4 & 73.6 & 51.7 & 61.9 & 57.9 & 55.3 & 56.7 & 76.5 \\
    HDMNet (CVPR'23) \cite{peng2023hierarchical} & 43.8 & 55.3 & 51.6 & 49.4 & 50.0 & 72.2 & 50.6 & 61.6 & 55.7 & 56.0 & 56.0 & 77.7 \\
    AMNet (NeurIPS'23) \cite{wang2023focus} & 44.9 & 55.8 & 52.7 & 50.6 & 51.0 & 72.9 & 52.0 & 61.9 & 57.4 & 57.9 & 57.3 & 78.8 \\
    HMNet (NeurIPS'24) \cite{xu2024hybrid} & 45.5 & 58.7 & 52.9 & 51.4 & 52.1 & 74.5 & 53.4 & 64.6 & 60.8 & 56.8 & 58.9 & 77.6 \\
    \midrule
    \addlinespace[2pt]
    \multicolumn{13}{l}{\textit{\textbf{Foundation-based FSS Methods}}} \\ 
    \addlinespace[3pt]
    Matcher (ICLR'24) \cite{liu2023matcher} & 52.7 & 53.5 & 52.6 & 52.1 & 52.7 & -- & 60.1 & 62.7 & 60.9 & 59.2 & 60.7 & -- \\
    VRP-SAM (CVPR'24) \cite{sun2024vrp} & 48.1 & 55.8 & 60.0 & 51.6 & 53.9 & -- & -- & -- & -- & -- & -- & -- \\
    GF-SAM (NeurIPS'24) \cite{zhang2024bridge} & 56.6 & 61.4 & 59.6 & 57.1 & 58.7 & -- & 67.1 & 69.4 & 66.0 & 64.8 & 66.8 & -- \\
    SANSA (NeurIPS'25) \cite{cuttano2025sansa} & 58.9 & 62.6 & 61.5 & 58.0 & 60.2 & -- & -- & -- & -- & -- & 67.3 & -- \\
    FSSAM (ICML'25) \cite{xu2025unlocking} & 59.9 & 65.6 & 62.1 & 61.6 & 62.3 & 77.3 & \underline{68.6} & \textbf{74.0} & 64.5 & 69.9 & 69.3 & 82.9 \\
    \midrule
    \addlinespace[2pt]
    \textbf{FSS-SAM3 (Ours)} & \underline{64.2} & \underline{66.8} & \underline{67.7} & \underline{67.5} & \underline{66.6} & \underline{82.9} & 68.6 & 72.5 & \underline{72.5} & \underline{73.6} & \underline{71.8} & \underline{84.9} \\
    \textbf{FSS-SAM3 + text (Ours)} & \textbf{74.0} & \textbf{75.0} & \textbf{76.7} & \textbf{77.5} & \textbf{75.8} & \textbf{87.1} & \textbf{74.5} & \underline{73.2} & \textbf{76.0} & \textbf{76.8} & \textbf{75.1} & \textbf{86.4} \\
    \bottomrule
  \end{tabular}
\end{table}

\subsection{Negative Prompt Failure}
\label{sec:neg_exp}

In practical scenarios, relying solely on positive prompts may be insufficient, especially when dealing with complex backgrounds or fine-grained visual distinctions. In principle, negative prompts are expected not only to help the model better separate the target class from visually similar or distracting regions, but also to provide more detailed guidance that reduces ambiguity in user intent. To investigate this possibility, we extend our framework by incorporating automatically generated negative exemplars. Specifically, we first perform an auxiliary segmentation step on the support image together with the positive prompt to obtain a set of candidate mask predictions. Predictions with confidence scores above 0.5 are treated as additional positive references, while those with lower confidence scores are regarded as negative references. These references are then converted into geometric prompts. Following the spatial collage strategy, the support and query images are combined into a unified canvas, with both positive and negative exemplars provided to SAM3 to generate the final segmentation results.

Contrary to our expectations, introducing negative exemplars leads to noticeable performance degradation in our setting. Both mIoU and FB-IoU decrease significantly compared to using only positive prompts. We further investigate the impact of the number of negative prompts given to the model by varying the number of negative exemplars (1, 3, and 5), as shown in Table~\ref{tab:exp_negnum}. The results show that performance deteriorates as the number of negative prompts increases, suggesting that SAM3 may be sensitive to negative guidance.

\begin{figure}[t]
     \centering
     \begin{subfigure}[b]{0.48\textwidth}
         \centering
         \includegraphics[width=0.49\linewidth, height=4cm]{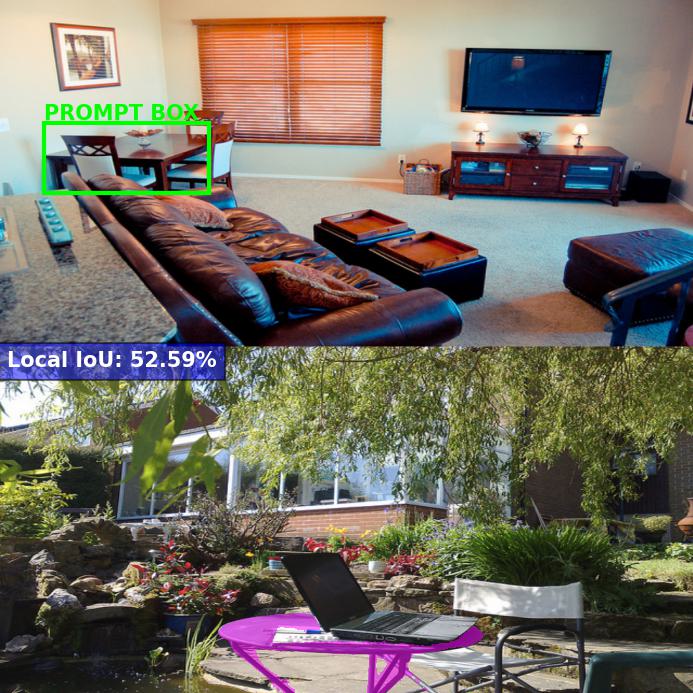}
         \includegraphics[width=0.49\linewidth, height=4cm]{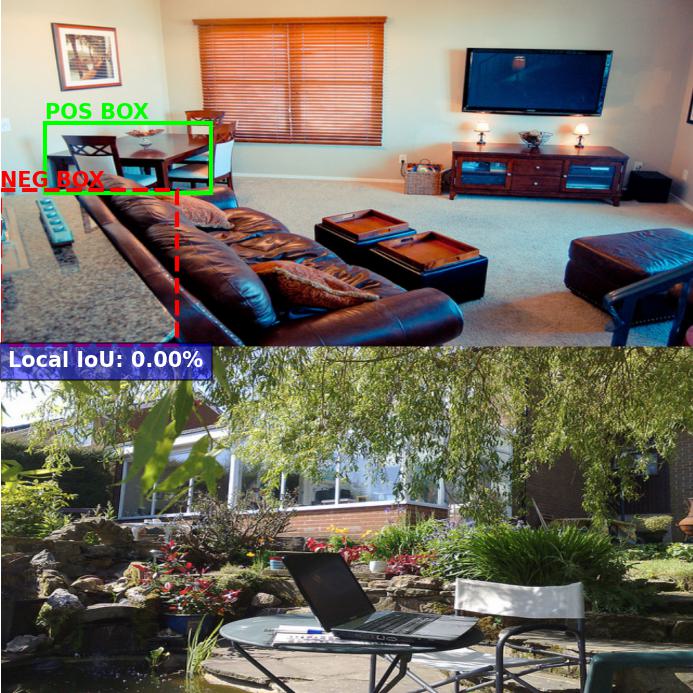}
         \caption{Comparison with and without negative prompts.}
         \label{fig:neg_comp}
     \end{subfigure}
     \hfill
     \begin{subfigure}[b]{0.48\textwidth}
         \centering
         \includegraphics[width=0.49\linewidth, height=4cm]{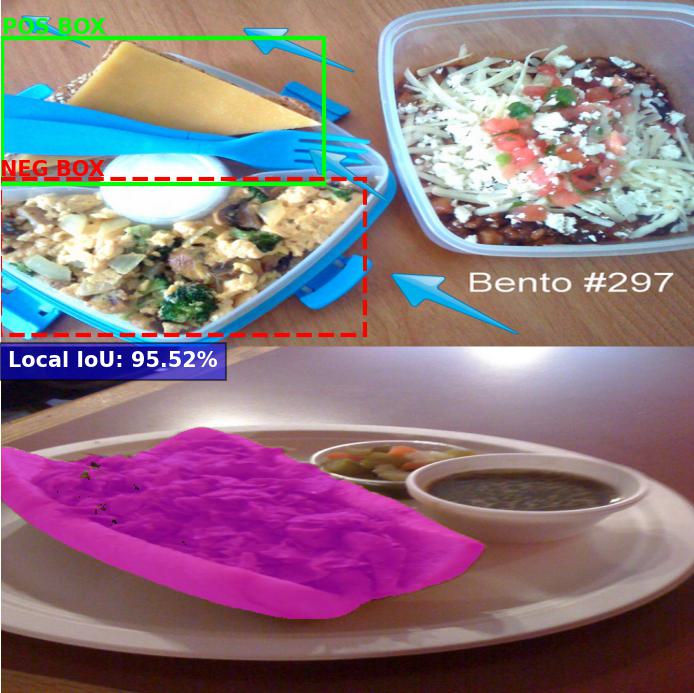}
         \includegraphics[width=0.49\linewidth, height=4cm]{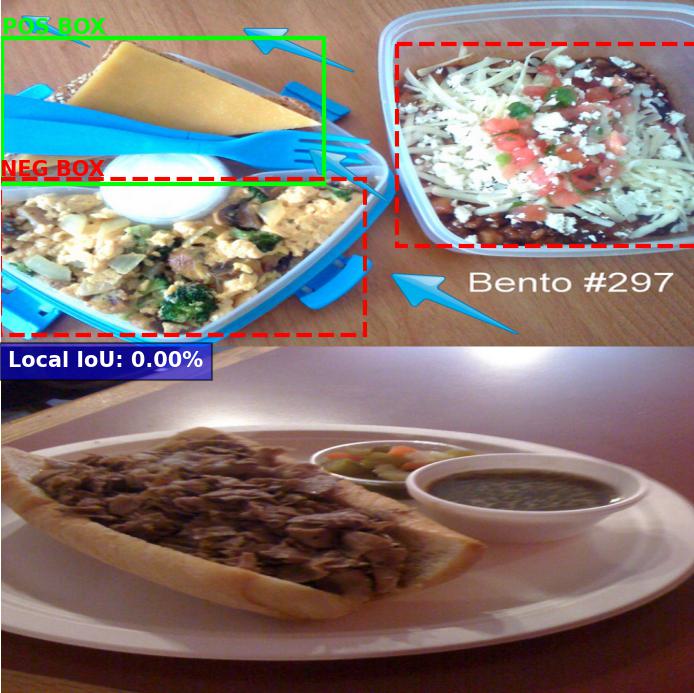}
         \caption{Effect of the number of negative prompts.}
         \label{fig:neg_num}
     \end{subfigure}
     
     \caption{Qualitative results of negative prompt interference in SAM3. (a) shows how adding a negative prompt can lead to prediction collapse where the target object is erroneously suppressed. (b) demonstrates the sensitivity to the number of negative prompts.}
     \label{fig:neg_vis}
\end{figure}

In addition to the quantitative results, we observe clear failure patterns in qualitative predictions. Negative prompts often lead to prediction collapse, where the model tends to classify most pixels as background and fails to produce meaningful segmentation masks (Figure \ref{fig:neg_vis}). One possible explanation is that the pretrained SAM3 is primarily optimized for positive prompt guidance and may not reliably interpret negative prompts when multiple regions are provided. Under our unified spatial canvas formulation, negative prompts may introduce conflicting spatial signals within the shared attention space, causing the model to over-suppress candidate regions. As a result, the model fails to properly localize the target object. To further analyze this phenomenon, we conduct additional controlled experiments, which are presented in Section~\ref{sec:neg_prompt}.

\begin{table}[t]
  \caption{Experiment on the number of negative prompts for COCO-$20^i$ dataset. $\Delta_m$ and $\Delta_F$ represent the decrease in mIoU and FB-IoU, respectively, compared to the $N_{neg}=0$ baseline.}
  \label{tab:exp_negnum}
  \centering
  \small
  \setlength{\tabcolsep}{3.5pt}
  \begin{tabular}{l @{\hspace{1.2em}} cccc cc cc}
    \toprule
    \multirow{2}{*}{Max Negative Prompts} & \multicolumn{4}{c}{COCO-$20^i$} & \multicolumn{2}{c}{mIoU} & \multicolumn{2}{c}{FB-IoU} \\
    \cmidrule(r){2-5} \cmidrule(lr){6-7} \cmidrule(l){8-9}
    & $20^0$ & $20^1$ & $20^2$ & $20^3$ & mIoU & $\Delta_m$ & FB-IoU & $\Delta_F$ \\
    \midrule
    $N_{neg} = 0$ (Positive Only) & 64.2 & 66.8 & 67.7 & 67.5 & 66.6 & -- & 82.9 & -- \\
    \midrule
    $N_{neg} \le 1$ & 51.1 & 57.5 & 54.2 & 56.6 & 54.8 & $-11.8$ & 77.2 & $-5.7$ \\
    $N_{neg} \le 3$ & 41.7 & 49.1 & 47.6 & 47.9 & 46.6 & $-20.0$ & 73.1 & $-9.8$ \\
    $N_{neg} \le 5$ & 40.0 & 48.1 & 45.9 & 44.9 & 44.7 & $-21.9$ & 72.4 & $-10.5$ \\
    \bottomrule
  \end{tabular}
\end{table}

\section{Ablation Study} 
In this section, we conduct a series of systematic ablation experiments on the PASCAL-$5^i$ and COCO-$20^i$ datasets to analyze the key design choices in our framework. Our analysis focuses on three aspects: spatial composition, multimodal prompting, and the behavior of negative prompts.

\begin{table}[t]
  \caption{Ablation study of spatial layout designs on PASCAL-$5^i$ and COCO-$20^i$. $S$ denotes the support images. "S Position" indicates the placement of $S$. FR: Forced Resizing; ARP: Aspect-Ratio Preserving. Mean mIoU and FB-IoU (\%) are reported.}
  \label{tab:ablation_res}
  \centering
  \small
  \setlength{\tabcolsep}{2pt}
\begin{tabular}{llc ccccc c ccccc c}
    \toprule
    \multirow{2}{*}{Layout Type} & \multirow{2}{*}{S Position} & \multirow{2}{*}{Ratio} & \multicolumn{6}{c}{PASCAL-$5^i$} & \multicolumn{6}{c}{COCO-$20^i$} \\
    \cmidrule(r){4-9} \cmidrule(l){10-15}
    & & & $5^0$ & $5^1$ & $5^2$ & $5^3$ & Mean & FB & $20^0$ & $20^1$ & $20^2$ & $20^3$ & Mean & FB \\
    \midrule
    \textit{1-Shot Strategy} & & & & & & & & & & & & & & \\
    ARP  & Top (Padded) & -- & 66.4 & 83.5 & 78.5 & 81.8 & 77.5 & 87.1 & 61.6 & 66.2 & 67.3 & 69.0 & 66.0 & 82.3 \\
    FR (Horizontal) & Left & 0.5 & 67.0 & \textbf{85.2} & 79.3 & 81.0 & 78.1 & 87.5 & 60.5 & 66.4 & 66.7 & 65.5 & 64.8 & 81.8 \\
    FR (Vertical)   & Bottom & 0.5 & 68.8 & 85.0 & 80.7 & 82.3 & 79.2 & 88.1 & 62.9 & \textbf{67.8} & 66.1 & \textbf{69.1} & 66.5 & 82.8 \\
    FR (Vertical)   & Bottom & 0.6 & 69.6 & \textbf{85.2} & 79.9 & 82.1 & 79.2 & 88.0 & \textbf{64.2} & 66.8 & \textbf{67.7} & 67.5 & \textbf{66.6} & \textbf{82.9} \\
    FR (Vertical)   & Top & 0.5 & 69.2 & 83.9 & 80.7 & 82.1 & 79.0 & 88.1 & 63.3 & 67.5 & 66.0 & 67.8 & 66.1 & 82.9 \\
    FR (Vertical)   & Top & 0.6 & \textbf{70.4} & 84.6 & \textbf{80.9} & \textbf{82.3} & \textbf{79.6} & \textbf{88.3} & 61.9 & 67.6 & 65.7 & 68.9 & 66.0 & 82.7 \\
    \midrule
    \textit{5-Shot Strategy} & & & & & & & & & & & & & & \\
    Uniform Grid     & $2 \times 3$ Grid & --   & 73.4 & 86.5 & 79.8 & 81.7 & 80.3 & 88.3 & 66.2 & 72.2 & 71.0 & 70.6 & 70.0 & 84.5  \\
    Vertical Strip   & Left & 0.4 & 72.6 & 86.5 & 79.6 & 82.9 & 80.4 & 88.3 & 67.5 & 71.3 & 70.0 & 72.6 & 70.4 & 84.0 \\
    Horizontal Strip & Top & 0.3 & 73.1 & 87.3 & 80.3 & \textbf{83.5} & 81.1 & 88.8 & 68.1 & 71.8 & 71.0 & 73.5 & 71.1 & 84.5 \\
    Inverse L-shape  & Top-Left & 0.4 & 74.3 & \textbf{87.5 }& \textbf{80.4} & 82.0 & 81.0 & 88.7 & 67.6 & 72.1 & \textbf{73.3} & 73.0 & 71.5 & 84.9 \\
    Inverse L-shape  & Top-Left & 0.3 & \textbf{74.8} & 86.8 & \textbf{80.4} & 82.9 & \textbf{81.2} & \textbf{88.9} & \textbf{68.6} & \textbf{72.5} & 72.5 & \textbf{73.6} & \textbf{71.8} & \textbf{84.9} \\
    \bottomrule
  \end{tabular}
\end{table}

\subsection{Spatial Design: Canvas Composition and Layout}
We first investigate how spatial composition affects model performance. Since the image encoder of SAM3 operates on fixed-resolution inputs, the arrangement and scaling of support and query images directly influence the attention distribution.
\paragraph{Composition strategy}
For the 1-shot setting, we compare Aspect-Ratio Preserving (ARP) and Forced Resizing (FR). ARP maintains the original aspect ratio with padding, while FR resizes all images to a fixed $1008 \times 1008$ canvas. As shown in Table~\ref{tab:ablation_res}, FR consistently outperforms ARP, indicating that maximizing token density within the attention space is more beneficial than preserving geometric fidelity. We further analyze the effects of different concatenation directions and spatial allocation ratios. Experimental results show that vertical concatenation (stacking images top and bottom) performs slightly better than other configurations. Moreover, the best performance is achieved when the support-to-query ratio is approximately 6:4.

\paragraph{Multi-shot layout}
In the multi-shot setting (5-shot), we further explore how to arrange multiple support images within the canvas. We compare four strategies:  
(1) \textit{Uniform grid}, where five support images and one query image are resized and arranged into a $2 \times 3$ grid;  
(2) \textit{Horizontal stacking}, where five support images are arranged in a row along the top of the canvas, and the query image occupies a larger region below;  
(3) \textit{Vertical stacking}, where five support images are arranged along the left side of the canvas, and the query image occupies a larger region on the right; and  
(4) \textit{Asymmetric inverse L-shape}, where support images are placed along the top and left boundaries, while the query image occupies a larger bottom-right region. Experimental results in Table~\ref{tab:ablation_res} show that the inverse L-shape significantly outperforms other layouts. Notably, compared with stacking-based strategies, the relative spatial allocation between support and query images is similar. Therefore, the performance difference mainly arises from the arrangement of support images rather than their size. This suggests that the spatial relationship between support and query images affects feature interactions within the shared attention space. Distributing support images along the canvas boundaries may reduce interference with the query region, thereby improving segmentation performance.

\subsection{Multimodal Prompting: Vision vs. Text}
\label{sec:multimodal}
We analyze the contribution of different modalities by comparing three settings: (1) using only positive visual prompts, (2) using only textual prompts (category names), and (3) combining both modalities. As shown in Table~\ref{tab:ablation_prompts}, the hybrid setting consistently achieves the best performance, indicating that visual and textual modalities are complementary. Specifically, visual exemplars provide instance-level appearance cues, while textual prompts offer category-level semantic guidance. We further investigate the scope of textual priors by varying the candidate label set, including ground-truth names, fold-level candidates, and dataset-level candidates. For each episode, we select the most relevant label from the candidate set based on the support image and use it as the textual prompt, which is then combined with the visual prompt to guide segmentation on the query image. As shown in Table~\ref{tab:ablation_text}, performance slightly degrades as the number of candidate labels increases, while remaining relatively stable overall. This indicates that even when the correct label is not explicitly provided, the model can still retrieve useful semantic cues from a broader candidate pool.

\begin{table}[t]
  \caption{Ablation study of different prompting strategies on COCO-$20^i$ and PASCAL-$5^i$ datasets. We report the Mean IoU (\%) and FB-IoU (\%) for 1-shot segmentation.}
  \label{tab:ablation_prompts}
  \centering
  \small
  \setlength{\tabcolsep}{2.5pt}
  \begin{tabular}{l @{\hspace{1.5em}} ccccc c ccccc c}
    \toprule
    \multirow{2}{*}{Prompting Strategy} & \multicolumn{6}{c}{PASCAL-$5^i$} & \multicolumn{6}{c}{COCO-$20^i$} \\
    \cmidrule(r){2-7} \cmidrule(l){8-13}
    & $5^0$ & $5^1$ & $5^2$ & $5^3$ & Mean & FB-IoU & $20^0$ & $20^1$ & $20^2$ & $20^3$ & Mean & FB-IoU \\
    \midrule
    Positive Visual Prompt & 70.4 & 84.6 & 80.9 & 82.3 & 79.6 & 88.3 & 64.2 & 66.8 & 67.7 & 67.5 & 66.6 & 82.9 \\
    Text Prompt            & 76.0 & 87.9 & 81.5 & 78.7 & 81.0 & 89.3 & 72.8 & 73.8 & 75.2 & 77.9 & 74.9 & 86.3 \\
    \midrule
    Both                   & \textbf{76.2} & \textbf{87.5} & \textbf{81.6} & \textbf{79.5} & \textbf{81.2} & \textbf{89.4} & \textbf{74.0} & \textbf{75.0} & \textbf{76.7} & \textbf{77.5} & \textbf{75.8} & \textbf{87.1} \\
    \bottomrule
  \end{tabular}
\end{table}

\begin{table}[t]
  \caption{Ablation study on the scope of textual priors for PASCAL-$5^i$ and COCO-$20^i$ datasets. We compare the use of ground-truth names against candidate labels selected from fold-level and dataset-level pools. Mean mIoU and FB-IoU (\%) are reported.}
  \label{tab:ablation_text}
  \centering
  \small
  \setlength{\tabcolsep}{2.5pt}
  \begin{tabular}{l @{\hspace{1.5em}} ccccc c ccccc c}
    \toprule
    \multirow{2}{*}{Scope of Textual Prior} & \multicolumn{6}{c}{PASCAL-$5^i$} & \multicolumn{6}{c}{COCO-$20^i$} \\
    \cmidrule(r){2-7} \cmidrule(l){8-13}
    & $5^0$ & $5^1$ & $5^2$ & $5^3$ & Mean & FB-IoU & $20^0$ & $20^1$ & $20^2$ & $20^3$ & Mean & FB-IoU \\
    \midrule
    Ground-truth Name         & \textbf{76.2} & \textbf{87.5} & \textbf{81.6} & \textbf{79.5} & \textbf{81.2} & \textbf{89.4} & \textbf{74.0} & \textbf{75.0} & \textbf{76.7} & \textbf{77.5} & \textbf{75.8} & \textbf{87.1} \\
    \midrule
    Fold-level Candidates     & 76.1 & 87.1 & 80.7 & 79.2 & 80.8 & 89.2 & 71.5 & 72.8 & 70.8 & 74.5 & 72.4 & 86.1 \\
    Dataset-level Candidates  & 76.1 & 84.6 & 80.2 & 75.6 & 79.1 & 88.4 & 66.0 & 69.1 & 70.0 & 66.3 & 67.8 & 83.2 \\
    \bottomrule
  \end{tabular}
\end{table}

\subsection{Limitations of Negative Prompting}
\label{sec:neg_prompt}
As discussed in Section~\ref{sec:neg_exp}, introducing negative prompts leads to performance degradation. To further investigate this phenomenon, we construct an additional evaluation setting on the MSCOCO dataset for analyzing negative prompting behavior. Unlike standard FSS benchmarks, we ensure that both support and query images contain at least two object categories, where one category is treated as the target (positive) and the others serve as sources for negative prompts. We consider three types of negative prompting: (1) \textit{background negatives}, where negative prompts are sampled from non-target regions such as sky or grass; (2) \textit{semantic distractors}, where another object category within the same image is used as a negative prompt; and (3) \textit{multiple negatives}, where all non-target objects in the image are used as negative prompts (up to 10 regions).

As shown in Table~\ref{tab:negative_analysis}, all three settings result in consistent performance degradation. The performance gap is particularly pronounced in the 1-shot setting, where mIoU drops by up to $27.2\%$, suggesting that additional support samples may provide a degree of regularizing effect against negative prompt interference. Notably, even background-based negative prompts adversely affect segmentation performance, suggesting that SAM3 may not reliably interpret negative guidance during inference. These findings indicate that the current SAM3 prompting mechanism struggles to balance positive and negative signals in few-shot settings. Rather than selectively suppressing irrelevant regions, negative prompts tend to broadly attenuate feature responses, weakening target representations and leading to prediction collapse.

\begin{table}[t]
  \caption{Robustness analysis of SAM3 against various negative prompting scenarios on the MSCOCO dataset. We compare the performance of positive prompts only against settings with semantic distractors, background negatives, and multiple negative regions. $\Delta$ indicates the absolute decrease in Mean IoU (\%).}
  \label{tab:negative_analysis}
  \centering
  \small
  \setlength{\tabcolsep}{10pt} 
  \begin{tabular}{l cc cc}
    \toprule
    \multirow{2}{*}{Prompting Scenario} & \multicolumn{2}{c}{1-shot} & \multicolumn{2}{c}{5-shot} \\
    \cmidrule(r){2-3} \cmidrule(l){4-5}
    & mIoU & $\Delta$ & mIoU & $\Delta$ \\
    \midrule
    Positive Only (Baseline)      & 53.2 & --      & 65.0 & -- \\
    \midrule
    Semantic Distractors          & 26.0 & $-27.2$ & 60.8 & $-4.2$ \\
    Background Negatives          & 22.1 & $-31.1$ & 61.4 & $-3.7$ \\
    Multiple Negatives ($\le 10$) & 9.7  & $-43.4$ & 52.6 & $-12.4$ \\
    \bottomrule
  \end{tabular}
\end{table}

\section{Conclusion}
In this work, we revisit Few-Shot Semantic Segmentation from the perspective of visual foundation models. We demonstrate that, by leveraging the Promptable Concept Segmentation capability of SAM3, it is possible to achieve strong FSS performance without any training, fine-tuning, or architectural modification. 

We propose a simple yet effective spatial concatenation strategy that transforms cross-image correspondence into a unified spatial reasoning problem. By embedding support and query images into a shared canvas, SAM3 is able to implicitly associate semantically related regions through its native attention mechanism. This formulation eliminates the need for explicit matching modules and establishes a strong training-free baseline for FSS.

In addition, we conduct a systematic analysis of prompt interactions and uncover a critical limitation: negative prompts, although theoretically beneficial, lead to consistent performance degradation in few-shot settings. Our findings suggest that current foundation models struggle to balance positive and negative signals, revealing an important direction for future research on prompt design and multimodal interaction.

Overall, our results highlight the potential of leveraging foundation models for few-shot segmentation through simple spatial formulations, while also emphasizing the need for a deeper understanding of prompt behavior in complex semantic tasks.

\section{Limitation}

Despite the strong performance of our training-free formulation, several limitations remain. First, as analyzed in Section~\ref{sec:neg_prompt}, the integration of negative prompts result in noticeable performance degradation in few-shot scenarios. Instead of effectively filtering distractors, negative prompts often trigger over-suppression, causing the collapse of correct prediction masks. This suggests that SAM3 lacks a robust mechanism to balance competing semantic signals when both positive and negative prompts are provided, particularly when operating under the constraints of limited supervision.

Second, although SAM3 supports both visual and textual inputs, we observe that its image and text encoders are not aligned within a shared semantic space. In exploratory experiments, we directly computed similarity between features from the support image and candidate class names for category matching. However, this approach yields suboptimal results, confirming that cross-modal similarity is inherently unreliable in this architecture.This is likely because image-text interaction is deferred to the decoder stage rather than being explicitly aligned at the encoder level. Consequently, SAM3 cannot serve as a CLIP-like model for cross-modal retrieval.

Finally, we investigated a temporal formulation by treating support and query images as a pseudo-video sequence, where the support image acts as a preceding frame stored in memory to guide segmentation. However, this approach failed to produce meaningful results, indicating that SAM3’s memory mechanism is optimized for temporal consistency (tracking) rather than semantic correspondence across disjoint images. While prior SAM2-based works like FSSAM~\cite{xu2025unlocking} and SANSA~\cite{cuttano2025sansa} achieve success via adaptation, our findings underscore that a naive temporal application is insufficient, highlighting a critical gap between temporal memory and cross-image semantic reasoning.

\section{Future Work}

An important direction for future research is to address the instability of negative prompting in few-shot settings, particularly for achieving precise, attribute-level control. While SAM3 can identify complex positive compositions through text prompts (e.g., "spotted dog"), it struggles with negation and the inherent ambiguity of visual-only prompting. Our ablations in Section\ref{sec:neg_prompt} show that negative prompts often cause over-suppression and even diminish target regions. Future work may explore prompt reweighting, positive-negative decoupling, or learning-based adaptation to suppress local distractors without weakening the global target representation, thereby improving robustness and enabling more fine-grained segmentation.

Furthermore, strengthening multimodal alignment within SAM3 remains a high-priority direction. Although textual and visual prompts currently offer complementary benefits at the decoder stage, the absence of encoder-level alignment limits the model's potential. Future work could explore lightweight alignment modules or contrastive learning objectives to bridge this gap. Enhanced alignment would facilitate more reliable cross-modal reasoning and a more seamless integration of visual exemplars with textual priors.

Lastly, given that spatial composition proved more effective than naive temporal formulations in FSS settings, future studies should investigate how to explicitly embed cross-image semantic correspondence into the memory architecture. This could involve adapting temporal attention mechanisms to prioritize instance-level matching over smooth motion tracking. More broadly, our study emphasizes that task formulation is as crucial as model architecture; hence, designing frameworks that better align with a foundation model's intrinsic properties will be key to unlocking its full potential in few-shot learning.

\bibliographystyle{plain}

\end{document}